\pdfoutput=1

\documentclass[11pt]{article}

\usepackage[preprint]{acl}

\usepackage{times}
\usepackage{latexsym}

\usepackage[T1]{fontenc}

\usepackage[utf8]{inputenc}

\usepackage{microtype}

\usepackage{inconsolata}
\usepackage[utf8]{inputenc}

\usepackage{amsthm,amssymb,amsmath,mathtools}
\usepackage{cleveref}
\crefname{section}{§}{§§}
\Crefname{section}{§}{§§}
\usepackage{xspace}
\mathtoolsset{showonlyrefs=true}
\usepackage{enumitem}
\usepackage{dsfont}
\usepackage{booktabs}
\usepackage{color}
\usepackage{multirow}
\usepackage{caption}
\usepackage{subcaption}
\usepackage{pgf, tikz, adjustbox}
\usepackage{tcolorbox}
\usepackage{xcolor}
\usepackage{float}
\usepackage{hyperref}

\title{Beyond Isolated Capabilities: Bridging Long CoT Reasoning and Long-Context Understanding}

\author{
    Yifei Wang, Linjing Li , Daniel Zeng\\
    $^1$ State Key Laboratory of Multimodal Artificial Intelligence Systems, \\
    Institute of Automation, Chinese Academy of Sciences, Beijing, China \\
$^2$ School of Artificial Intelligence, University of Chinese Academy of Sciences, Beijing, China \\
\texttt{\{wangyifei2022,  linjing.li, dajun.zeng\}@ia.ac.cn} \\
}

\begin{document}
\maketitle
\begin{abstract}
Reasoning distillation has emerged as an effective approach to enhance the reasoning capabilities of smaller language models. However, the impact of large-scale reasoning distillation on other critical abilities, particularly in-context retrieval and reasoning, remains unexplored. This gap in understanding is particularly significant given the increasing importance of Retrieval-Augmented Generation (RAG) systems, where efficient acquisition and utilization of contextual information are paramount for generating reliable responses.
Motivated by the need to understand how the extended long-CoT process influences long-context comprehension, we conduct a comprehensive investigation using a series of open-source models distilled from Deepseek-R1, renowned for its exceptional reasoning capabilities. Our study focuses on evaluating these models' performance in extracting and integrating relevant information from extended contexts through multi-document question and answering tasks.
Through rigorous experimentation, we demonstrate that distilled reasoning patterns significantly improve long-context understanding. Our analysis reveals that distillation fosters greater long-context awareness by promoting more detailed and explicit reasoning processes during context analysis and information parsing. This advancement effectively mitigates the persistent "lost in the middle" issue that has hindered long-context models. 
\end{abstract}

\section{Introduction}
\begin{figure}[!ht]
    \centering
    \includegraphics[width=\linewidth]{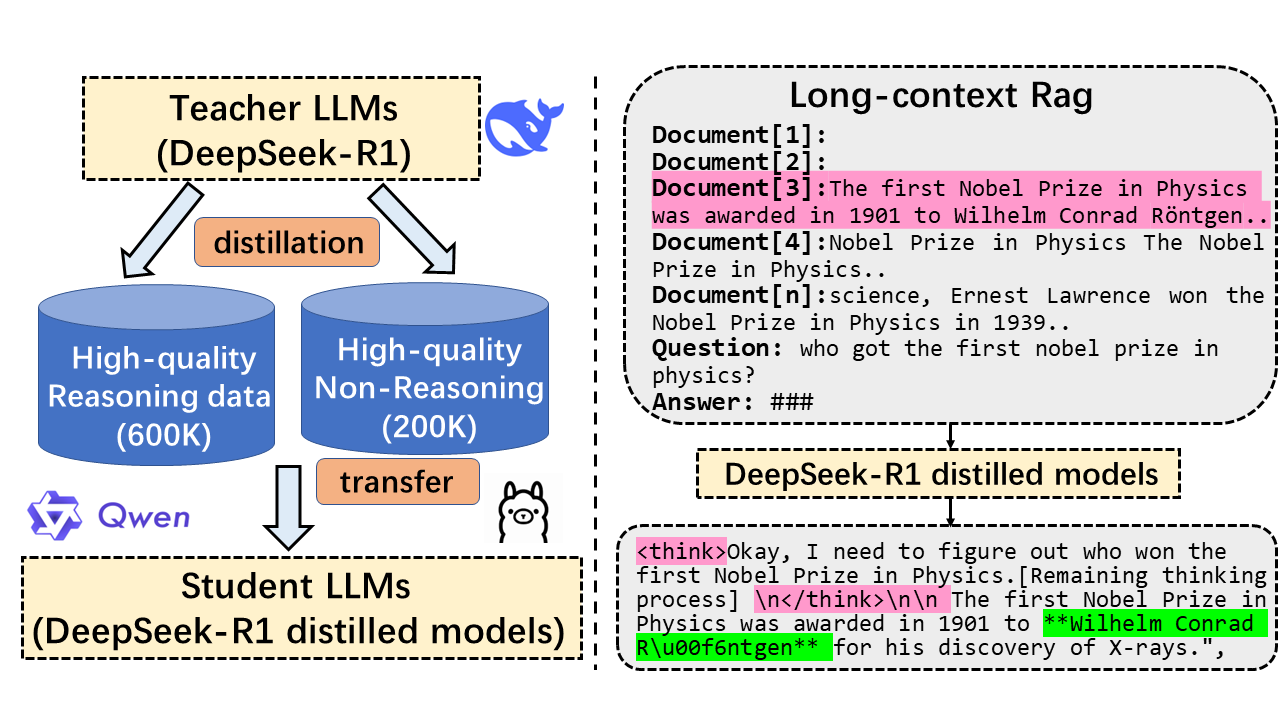}
    \caption{The left panel illustrates the distillation process from DeepSeek-R1, wherein the distilled LLMs acquire the capability to emulate advanced reasoning patterns, such as producing extended CoT through iterative reflection and verification cycles, which is similarly manifested in the model's contextual comprehension, including thinking process in special token tag <think> and arrive the final response.}
    \label{fig:1}
    \vspace{-.4cm}
\end{figure}

Reasoning capability constitutes a fundamental pillar of large language models (LLMs), and researchers are continually pushing the boundaries to explore its upper limits across domains such as mathematics~\citep{li2024gsm,ahn2024large,luo2024reasoning}, coding~\citep{gu2401cruxeval}, and logical reasoning~\citep{pan2023logic}. Recent breakthroughs have been primarily driven by two complementary approaches. The first, exemplified by OpenAI's o1 series models~\citep{openai_o1_2024}, has achieved remarkable performance on complex reasoning benchmarks through inference-time scaling~\citep{snell2024scalingllmtesttimecompute} via innovative extensions of Chain-of-Thought (CoT) reasoning processes~\citep{wei2023chainofthoughtpromptingelicitsreasoning,qin2024o1replicationjourneystrategic,wang2025drtdeepreasoningtranslation,yeo2025demystifyinglongchainofthoughtreasoning,pang2025boltbootstraplongchainofthought}. In parallel, DeepSeek has introduced a groundbreaking paradigm through DeepSeek-R1 series~\citep{deepseekai2025deepseekr1incentivizingreasoningcapability}, leveraging large-scale reinforcement learning (RL) frameworks~\citep{shao2024deepseekmathpushinglimitsmathematical,luong2024reftreasoningreinforcedfinetuning} to substantially advance reasoning performance. 
A pivotal contribution of DeepSeek is its successful transfer of advanced reasoning capabilities to smaller-scale LLMs in Tab.\ref{tab1} through systematic distillation of reasoning patterns from the highly capable DeepSeek-R1 model (Fig.\ref{fig:1})\footnote{\url{https://huggingface.co/deepseek-ai/DeepSeek-R1-Distill-Llama-8B}}. This breakthrough enables resource-efficient models to emulate sophisticated reasoning patterns, thereby democratizing access to high-performance reasoning systems.

This advancement raises a critical question: \textbf{How does intensive distillation affect the in-context retrieval and reasoning capabilities of long-context LLMs?} In other words, the impact of reasoning distillation on the model's contextual understanding remains unclear and under-explored. While reasoning distillation—a technique that trains smaller models using extensive high-quality reasoning data—is effective for task-specific reasoning, it may inadvertently limit the fundamental capabilities of long-context models, such as the ability to comprehend and process extensive inputs. Conversely, enhancements in reasoning capabilities for tasks like mathematics and coding do not necessarily lead to improved utilization of contextual information in tasks requiring long-context RAG~\citep{leng2024long}.

The preservation of long-context awareness during distillation is crucial for two primary reasons. First, modern LLM systems are increasingly required to process extensive information within their expanded context windows ~\citep{bai-etal-2024-longalign, merth2024superpositionpromptingimprovingaccelerating, 10.5555/3692070.3692785}, offering an alternative to traditional post-training paradigms~\citep{xiao2023smoothquant}. Second, the growing prominence of RAG systems~\citep{gao2024retrievalaugmentedgenerationlargelanguage, lewis2020retrieval} underscores the need to address the inevitable obsolescence of parametric knowledge~\citep{cohen-etal-2024-evaluating, meng2022locating, onoe-etal-2023-lms}. The effectiveness of these RAG systems~\citep{ma2025blockattention, jin2024ragcacheefficientknowledgecaching} hinges on a model's ability to efficiently process and synthesize information~\citep{qiu2025elicitingincontextretrievalreasoning} within large context windows~\citep{liu-etal-2024-lost, ivgi-etal-2023-efficient, yue2025inference}.

To investigate the impact of acquiring advantageous thought patterns on contextual retrieval and reasoning in long-context scenarios, we conduct a systematic empirical evaluation of distilled models using tasks that necessitate accessing and utilizing information from extensive inputs. Specifically, we employ the multi-document question answering (MDQA) task \citep{liu-etal-2024-lost,baker2024lostmiddleinbetweenenhancing,tang2024multihoprag}, which are utilized to evaluate models' capabilities in retrieving pertinent information from provided documents and applying reasoning to answer given questions. To ensure a thorough and precise comparison of context understanding, we systematically vary context lengths and manipulate the position of relevant information within the input, while concurrently considering other potential factors such as the influence of parametric knowledge and common CoT patterns.

\begin{table}[!ht]
\centering
\small
\begin{tabular}{lp{2cm}@{\hspace{-0.3cm}}c}
\toprule
\textbf{Base Model} & \textbf{ Distilled} & \textbf{Support} \\
\midrule
\midrule
Qwen2.5-14B & \checkmark & 128K \\
Qwen2.5-32B &\checkmark & 128K \\
Llama-3.1-8B & \checkmark & 128K \\
Llama-3.3-70B-Instruct & \checkmark & 128K \\
\bottomrule
\end{tabular}
\caption{The corresponding base models which have distilled versions from DeepSeek-R1. Their distilled counterparts demonstrates exceptional reasoning capabilities, particularly in challenging tasks such as AIME 2024 and MATH-500, as evidenced by \cite{deepseekai2025deepseekr1incentivizingreasoningcapability}. \emph{All base LLMs used for distillation are equipped with long-context capabilities}.}
\label{tab1}
\vspace{-.5cm}
\end{table}

\section{Impact of Distilled Reasoning Patterns on Long-Context Understanding }
\begin{table*}[!t]
    \centering
    \small
    \vspace{-2ex}

    \resizebox{0.95\textwidth}{!}{
    \begin{tabular}{p{3.9cm}@{\hspace{-0.3cm}}ccccccc}
    \toprule
        \multirow{2}{*}{\bf Methods} &  \multirow{1}{*}{\bf Closed-} &  \multicolumn{6}{@{}c}{{\bf Open-Book}} \\
        && 10-doc  &  20-doc & 30-doc & 50-doc & 80-doc & AVG. \\
    \midrule
    \textit{Llama-3.1-8B direct QA} &  31.16&54.87&53.34&51.26&49.48&47.13&51.21\\
    \textit{Llama-3.1-8B Zero-CoT} & 24.62&34.52{\scriptsize -20.35}&34.67{\scriptsize -18.67}&30.59{\scriptsize -20.67}&32.71{\scriptsize -16.77}&30.81{\scriptsize -16.32}&32.66{\scriptsize -18.56}\\
    \textit{Llama-3.1-8B distilled} &22.61&\bf 62.91{\scriptsize +8.04} &\bf 57.86{\scriptsize +4.52} &\bf 55.05{\scriptsize +3.79}&\bf 52.01{\scriptsize +2.53}&\bf 49.29{\scriptsize +2.16}&\bf 55.42{\scriptsize +4.21}\\
    \midrule
    \textit{Qwen2.5-14B direct QA}&
    38.19&68.14&61.41&55.67&52.21&48.15&57.12\\
    \textit{Qwen2.5-14B Zero-CoT} &
    44.72&71.06{\scriptsize +2.92}&63.94{\scriptsize +2.53}&55.56{\scriptsize -0.11}&54.12{\scriptsize +1.91}&51.03{\scriptsize +2.88}&59.14{\scriptsize +2.03} \\
    \textit{Qwen2.5-14B distilled} &30.65&	\bf70.52{\scriptsize +2.38}&\bf66.41{\scriptsize +5.00}&\bf63.04{\scriptsize +7.37}&\bf62.26{\scriptsize +10.05}&\bf61.44{\scriptsize +13.29}&\bf64.73{\scriptsize +7.62}\\
    \midrule
    \textit{Qwen2.5-32B direct QA} & 40.20&	73.62&67.21&65.10&57.99&55.50&63.88\\
    \textit{Qwen2.5-32B Zero-CoT} &43.21&71.81{\scriptsize -1.81}&64.45{\scriptsize -2.76}&62.87{\scriptsize -2.23}&55.12{\scriptsize -2.87}&50.11{\scriptsize -5.39}&60.87{\scriptsize -3.01} \\
    \textit{Qwen2.5-32B distilled} &40.70&\bf 73.42{\scriptsize -0.02} &\bf 69.89{\scriptsize +2.68}&\bf 68.39{\scriptsize +3.29}&\bf 65.28{\scriptsize +7.29}&\bf 64.96{\scriptsize +9.46}&\bf 68.39{\scriptsize +4.50} \\
    \midrule
    \textit{Llama-3.3-70B direct QA} & 67.42&83.37&80.08&76.78&73.28&71.51&77.01\\
    \textit{Llama-3.3-70B Zero-CoT} 
    &65.66&80.58{\scriptsize -2.79}&76.88{\scriptsize -3.19}&74.02{\scriptsize -2.76}&69.44{\scriptsize -3.84}&69.32{\scriptsize -2.19}&74.05{\scriptsize -2.19}
\\
    \textit{Llama-3.3-70B distilled} 
    &  56.13&\bf 83.41{\scriptsize +0.04}&\bf82.51{\scriptsize +2.43}&\bf79.56{\scriptsize +2.78}&\bf77.41{\scriptsize +4.13}&\bf 76.32{\scriptsize +4.81}&\bf 79.84{\scriptsize +2.82}
 \\
    \bottomrule
    \end{tabular}
    }

        \caption{ Comparative analysis of EM scores: baselines vs. Distilled Counterparts across context Lengths.
    }
        \label{tab:main_1}
        \vspace{-.4cm}
\end{table*}
\subsection{Long-Context Probing Task}
\paragraph{MDQA} provides a rigorous benchmark for assessing the efficacy of distilled models in extracting pivotal information and reasoning over extended contexts. Formally, the task comprises $\mathbf{N}$ documents (docs), denoted as $D=\{d_i\}_{i=1}^n$, which collectively form an external knowledge repository $\mathcal{K} = \bigcup_{i=1}^n d_i$. These docs are systematically organized into a structured prompt:
$T_{prompt} = [d_1 \oplus d_2 \oplus \cdots \oplus d_n \oplus x_{query}],$
where $\oplus$ denotes the concatenation operator. The evaluation protocol ensures that the target information $d_{gold}$ is embedded within $D$, while introducing highly relevant yet non-essential distractor docs $d_{distractors}$. $d_{distractors}$ is selected based on their semantic similarity to the query (i.e., $sim(d_{distractors}, x_{query}) > \tau$), as determined by the retrieval system.

    

\subsection{Experiments}
\paragraph{Task Setup}  
We follow \citet{liu-etal-2024-lost} for single-hop MDQA, where the answer $x_{query}$ can be directly retrieved from a single document $d_{gold}$, utilizing the NaturalQuestions dataset \citep{kwiatkowski-etal-2019-natural}. For each query, we select the top-$k$ documents, including one golden doc and $k-1$ distractors, with $k \in \{10,20,30,50,80\}$ to vary contextual retrieval difficulty. To eliminate the inaccuracy by the "Lost in the Middle" problem in long-context LLMs\footnote{Performance degradation when key information appears in the middle of the input context.}, we randomize  $d_{gold}$ across 10 equally spaced positions and report averaged Exact Match (EM) scores to ensure robust evaluation.

\paragraph{Models and Baselines} Our evaluation framework incorporates a comprehensive suite of DeepSeek-R1 series distilled models, spanning from 8B to 70B parameters, alongside their corresponding foundation LLMs, as systematically presented in Table \ref{tab1}. To comprehensively understand how distilled Reasoning Patterns impact the capacity of in-context retrieval and reasoning, we introduce several baselines below:

\emph{\textbf{Weak Baselines}}: 
1) Base models under the close-book setting: this configuration is designed to isolate and quantify the impact of parametric knowledge by excluding non-reasoning data from the distillation process as illustrated in Fig. \ref{fig:1}.

\emph{\textbf{Strong Baselines}}:
2) A direct QA for base LLMs under open-book setting;
3) Zero-CoT QA for base LLMs under open-book setting. This baseline enables a rigorous comparison between the self-instruct thought processes inherent in foundation models and the distilled thought patterns.
\begin{figure*}[!h]
    \centering
    \begin{subfigure}[b]{0.23\linewidth}
        \centering
        \includegraphics[width=\textwidth]{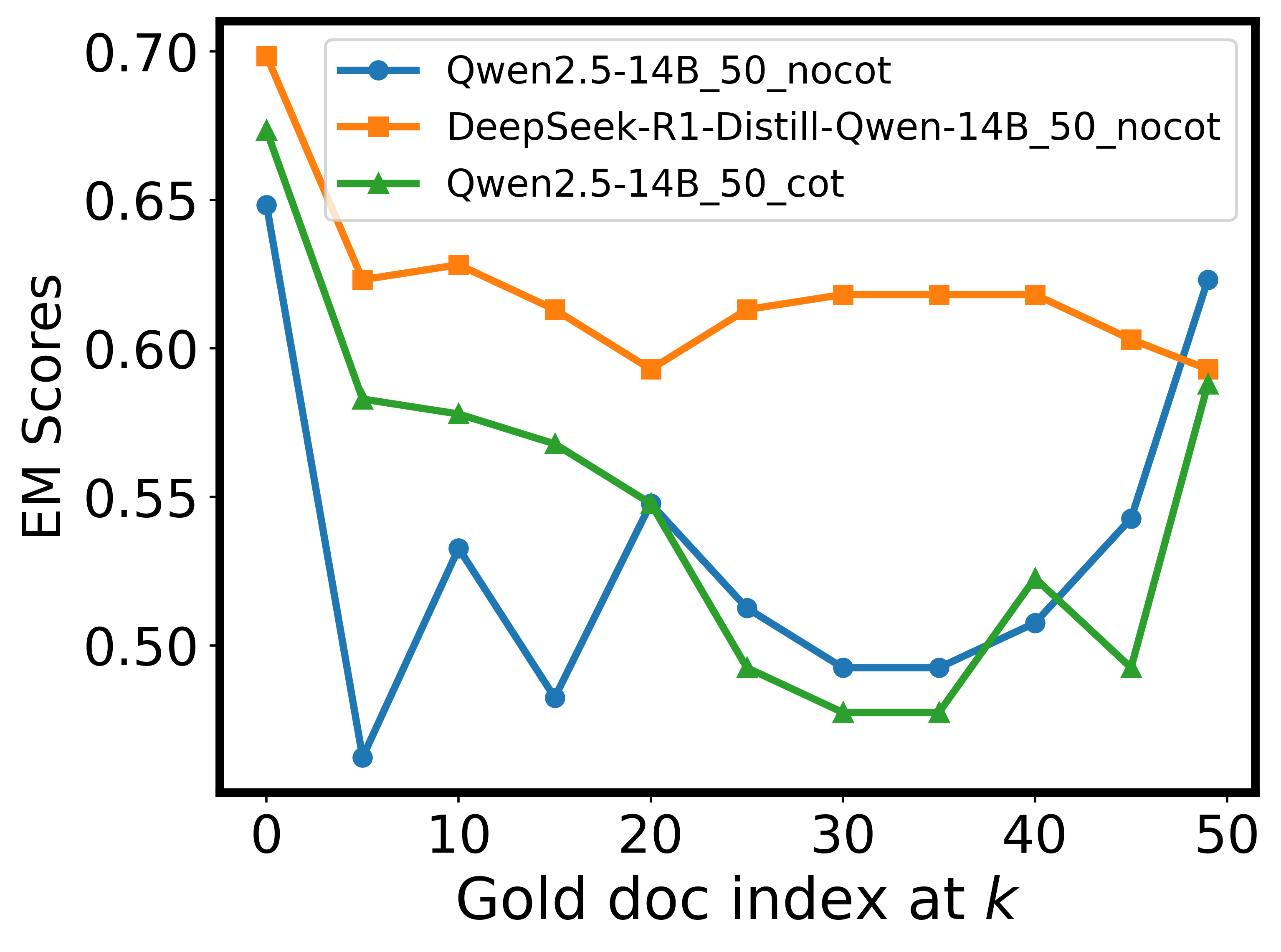}
        \label{fig:first}
    \end{subfigure}
    \hfill 
    \begin{subfigure}[b]{0.23\linewidth}
        \centering
        \includegraphics[width=\textwidth]{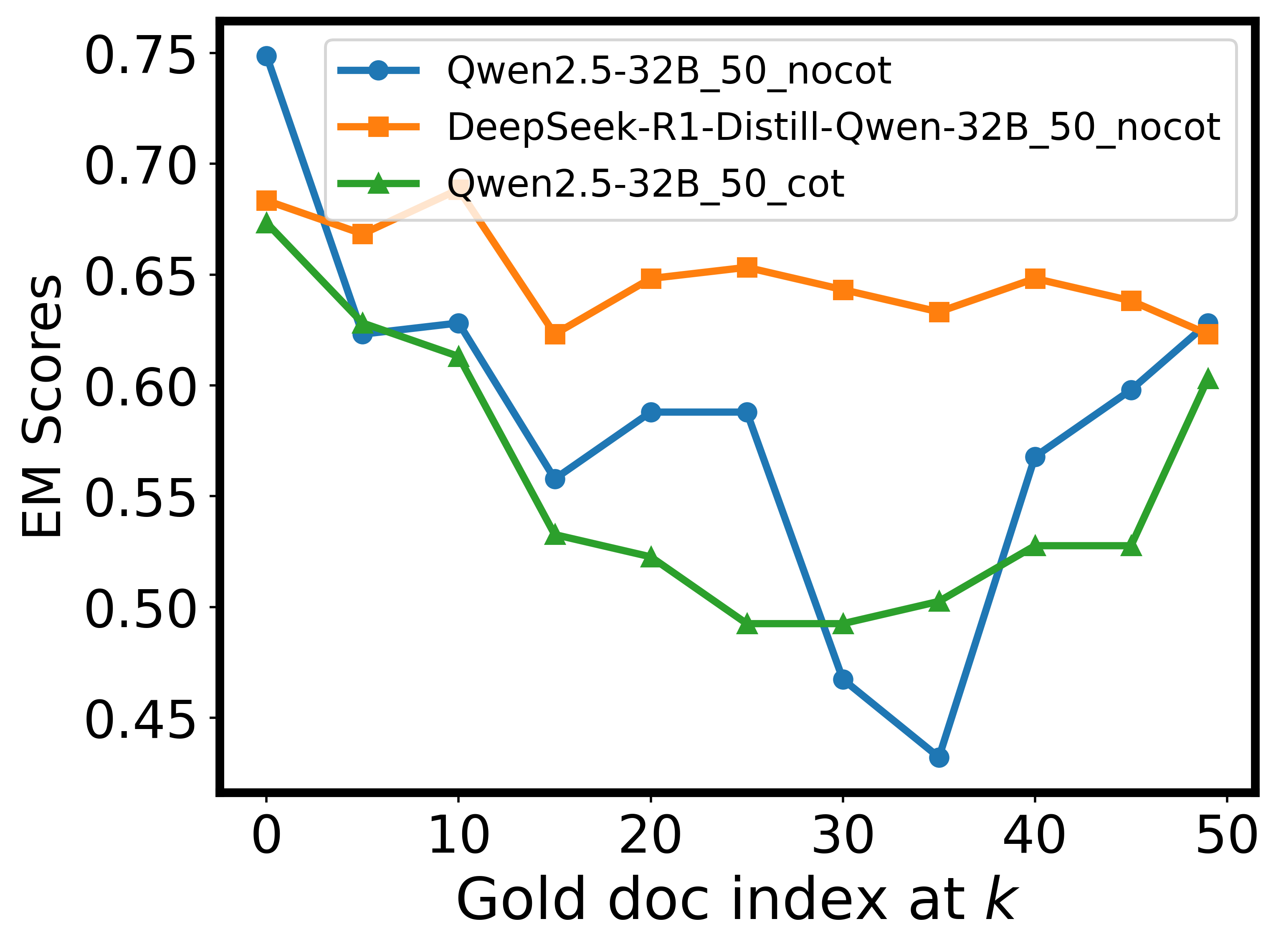}
        \label{fig:second}
    \end{subfigure}
    \hfill
    \begin{subfigure}[b]{0.23\linewidth}
        \centering
        \includegraphics[width=\textwidth]{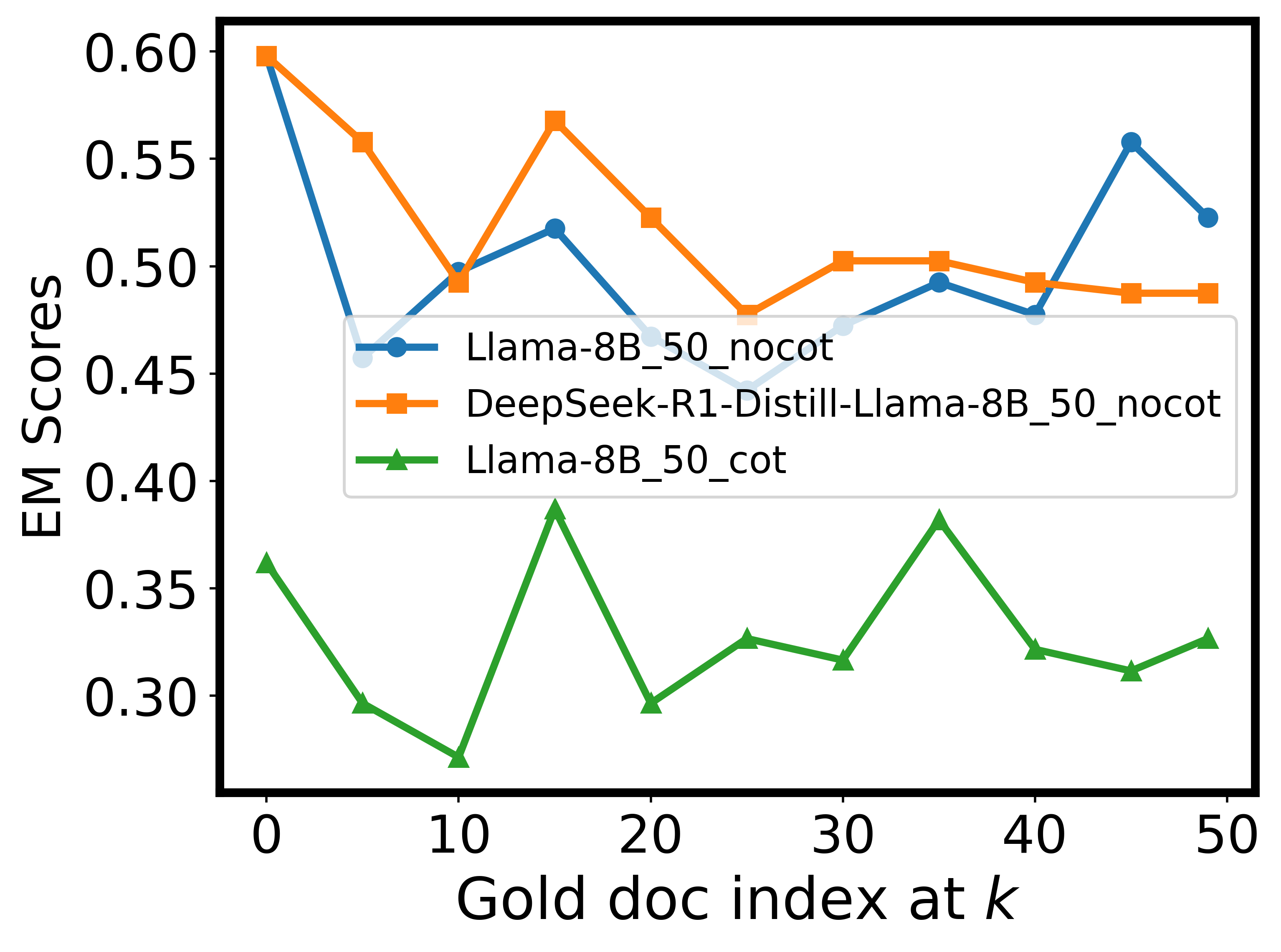}
        \label{fig:third}
    \end{subfigure}
    \hfill
    \begin{subfigure}[b]{0.23\linewidth}
        \centering
        \includegraphics[width=\textwidth]{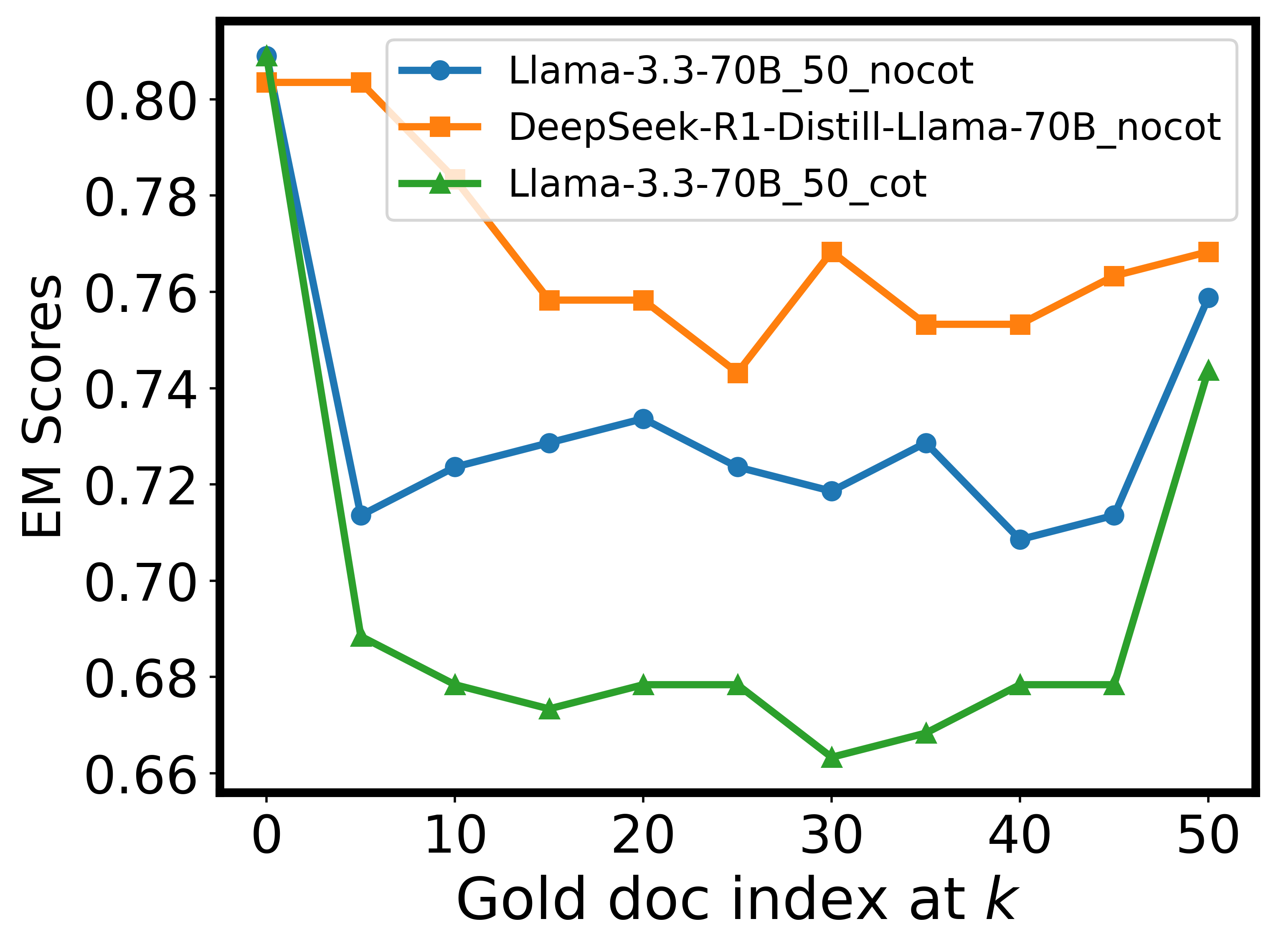}
        \label{fig:fourth}
    \end{subfigure}
    \vspace{-.7cm}
    \caption{Positional effects on EM scores in 50-doc QA.}

    \label{fig. position bias}
    \vspace{-.4cm}
\end{figure*}
\subsection{Main Results}
\paragraph{Implementation Details} 
In our experimental setup, we adopted the DeepSeek-R1 training template, used for the distillation process, as our prompt template for distilled models. To facilitate comprehensive CoT reasoning in distilled models, we configured the max\_new\_tokens parameter to 4096, ensuring sufficient capacity for extended reasoning sequences. Following established best practices, we implemented a temperature setting of 0.6 to maintain consistency and reduce repetitive patterns. For computational efficiency, all experiments were executed using the vLLM inference framework, executed on 8 NVIDIA A100 GPUs.

\paragraph{Long-context understanding.} As demonstrated in Tab.\ref{tab:main_1}, fine-tuning non-reasoning data during the distillation process does not significantly expand the parametric knowledge of foundational LLMs. This is evidenced by the notably lower accuracy of distilled LLMs under closed-book conditions compared to their base counterparts. However, the superior performance of distilled LLMs in open-book scenarios highlights their enhanced ability to identify and utilize pertinent information from extensive context windows to formulate accurate responses. Specifically, for Qwen2.5-14B and Qwen2.5-32B, the distilled versions exhibit the most significant improvements in in-context retrieval and reasoning, particularly as the context length increases. For instance, accuracy increases by 13.29 percent when the context includes 80 documents, underscoring the critical role of advanced thought patterns in managing long inputs.

\paragraph{Position bias.} Long-context LLMs in Fig.\ref{fig. position bias} suffer from a notable position bias, where performance greatly improves when \(d_{gold}\) is located at either extremity of the input sequence. This phenomenon can be attributed to two distinct mechanisms: the attention sink effect at the initial positions ~\citep{gu2025when} and recency bias at the terminal positions. In contrast, distilled models demonstrate remarkable positional invariance, maintaining consistent performance regardless of $d_{gold}$'s location. Particularly noteworthy is DeepSeek-R1-Distill-Qwen2.5-14B, which exhibits stable retrieval accuracy across a broad spectrum of doc positions (index 5-45), outperforming baseline models that show substantial performance fluctuations. This empirical evidence suggests that reasoning capacity effectively enhances the model's contextual retrieval capabilities, mitigating positional bias in multi-doc processing scenarios. Furthermore, the improved performance stability indicates enhanced awareness of the middle context window, mitigating a critical limitation in conventional long-context models.

\paragraph{Reasoning patterns.} The superior contextual awareness demonstrated by DeepSeek-like reasoning mechanisms stems from their incorporation of advanced cognitive processes, which significantly outperform traditional direct QA paradigms in MDQA tasks. To empirically validate the effectiveness of these reasoning patterns, we conducted a systematic comparison with zero-CoT, the default reasoning process in LLMs. Our experimental results, as presented in Table \ref{tab:main_1}, reveal that zero-CoT exhibits notable limitations in contextual utilization, particularly evident in its performance degradation across both Llama-3.1-8B and Qwen2.5-32B. These findings underscore the superiority of distilled reasoning patterns, as they facilitate sophisticated multi-source information processing, conflict resolution, and knowledge integration.

\subsection{Analysis and Discussion}
\begin{figure}[!ht]
    \centering
\includegraphics[width=\linewidth]{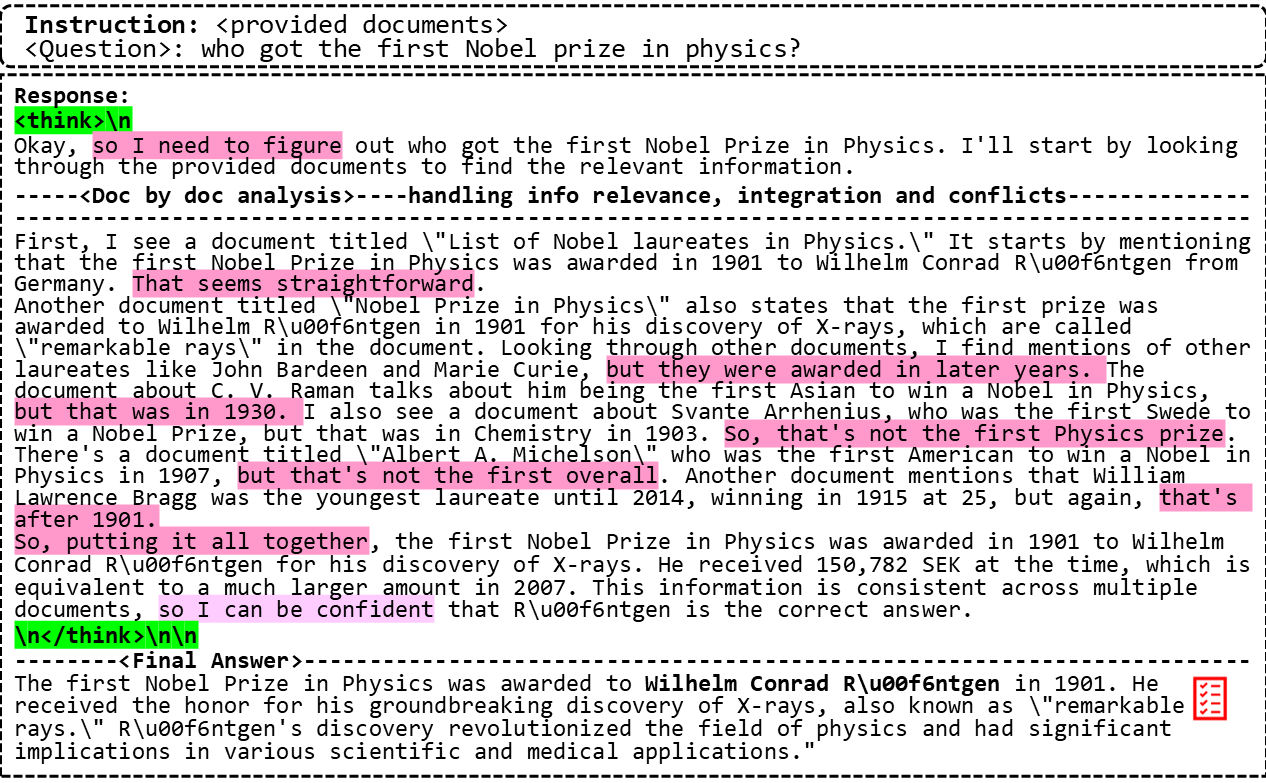}
        \caption{Reasoning pattern analysis of distilled models in long-context scenarios: a case study of DeepSeek-R1 on 50-doc QA. Refer to more instances in Appendix \ref{appendix:instances}.}
        \label{fig:patterns}
        \vspace{-.4cm}
\end{figure}

\paragraph{How distilled reasoning enhances long-context handling?}  
As exemplified in Fig.\ref{fig:patterns}. 
unlike base LLMs, which often struggle with middle-positioned docs and fail to recognize useful content, distilled LLMs employ a refined reasoning process with reflection and verification over multi docs.This extended thought process enables them to identify and discuss relevant documents that share the same topic or reference the same figures.
By conducting a doc-by-doc analysis, distilled LLMs meticulously examine the alignment between key details in the documents (e.g., figures, temporal references, or specific events) and the critical information in the query, allowing them to resolve inconsistencies and reconcile conflicts across multiple docs, ensuring a higher degree of accuracy. For instance, in the case of identifying the first Nobel Prize in Physics, the model not only extracts the relevant information about Wilhelm Conrad Röntgen but also cross-references other documents to verify the consistency of details such as the year of the award, the discovery of X-rays, and the monetary value of the prize. This thorough verification process minimizes the risk of errors and enhances the reliability of the information.
After completing the individual document analysis, distilled LLMs integrate all relevant information, synthesizing it into a coherent and accurate understanding of the long-context content. This systematic approach not only improves context awareness but also enhances the overall reliability of the information synthesis process. 
\paragraph{A broader View} Previous research has predominantly focused on enhancing long-context processing through innovative architectural modifications, particularly in redesigning position embedding and causal mask modules. Contrary to conventional approaches, our findings suggest that scaling inference-time reasoning significantly enhances the model's capacity to process extended inputs and effectively utilize contextual information. This paradigm shift offers a novel pathway for advancing long-context comprehension, independent of modifications to core architectural components.

\section{Conclusion}
\vspace{-.2cm}
Our research establishes a critical connection between two fundamental capabilities of contemporary models: reasoning proficiency and long-context comprehension. Through systematic investigation, we demonstrate that distilled reasoning patterns significantly enhance long-context understanding by enabling refined multi-document reasoning processes. This methodological advancement effectively mitigates key challenges in information consistency maintenance and cross-document integration, ultimately leading to more reliable and coherent response generation.


\newpage
\section*{Limitation}
While this study provides valuable insights into the reasoning patterns of distilled models in long-context scenarios, several limitations should be acknowledged. First, the evaluation is primarily conducted on single-hop MDQA tasks, which, although effective for assessing in-context retrieval, do not fully capture the complexities of multi-hop reasoning across disparate documents. Future work should extend the analysis to multi-hop scenarios to better understand the scalability of distilled reasoning patterns. Second, the study relies on the NaturalQuestions dataset, which may not fully represent the diversity of real-world long-context QA tasks. Incorporating additional datasets with varied domains and document structures could enhance the generalizability of the findings. Third, the positional bias analysis, while insightful, is limited to a fixed set of document positions (e.g., 10 equally spaced positions). A more granular investigation of positional effects, including varying document lengths and more nuanced position distributions, could provide deeper insights into the robustness of distilled models. 
\section*{Broader Impact}
Our work demonstrates that long CoT reasoning can significantly enhance models' ability to understand and utilize long-context information, offering a novel perspective beyond traditional approaches focused solely on architectural innovations or interpretability methods. By shifting the focus towards reasoning capabilities, this research opens new avenues for improving long-context understanding without relying exclusively on adaptions on core modules. This insight has the potential to democratize access to long-context AI systems, as it suggests that even smaller models, when equipped with robust reasoning mechanisms, can achieve competitive performance.  

Furthermore, our findings could impact domains that heavily rely on long-context comprehension, such as legal document analysis, medical research, and scientific literature review, by enabling more efficient and accurate information extraction. However, we also acknowledge potential risks, such as the misuse of enhanced reasoning capabilities for generating misleading or biased content. We advocate for ongoing research into ethical guidelines and safeguards to ensure that advancements in long-context understanding are aligned with societal values and deployed responsibly.  

This work encourages the community to rethink the priorities in long-context research, emphasizing reasoning as a complementary dimension to structural and interpretability improvements, ultimately contributing to more versatile and reliable AI systems.  

\newpage
\bibliography{cal_references}
\newpage
\onecolumn
\appendix

\section{Related Work}
\paragraph{Long-context LLMs} The field of long-context large language models (LLMs) has witnessed remarkable progress through three principal technical innovations: continued fine-tuning \cite{rozière2024codellamaopenfoundation}, position extrapolation \cite{10.1016/j.neucom.2023.127063}, and novel architectural designs \cite{peng-etal-2023-rwkv,gu2024mamba}. A significant portion of this research has focused on advancing positional embedding (PE) extrapolation techniques, with scholars developing systematic approaches to redesign PE methodologies for extended context windows \cite{zhu2024pose,xiao2024efficient,hetwo}.

Despite these advancements, a critical challenge has surfaced: the "lost in the middle" phenomenon identified by \cite{liu-etal-2024-lost}, where models struggle to effectively utilize information from the middle segments of their context windows. This limitation has spurred a new wave of research focused on optimizing middle-of-context retrieval within existing context lengths. Recent "found in the middle" approaches have emerged as promising solutions, targeting this specific weakness through innovations in core LLM components such as PE, causal masks, and inner hidden states \cite{chi-etal-2023-latent,lin2024mixture,peysakhovich2023attentionsortingcombatsrecency,wang2025eliminating,chen-etal-2024-fortify,zhang2024found}. Additionally, interpretable methods have been developed to mitigate position bias in long-context processing \cite{yu2025mitigate,an2024make}.
While reasoning and long-context understanding are fundamental to current LLM capabilities, the interplay between these two aspects remains underexplored. Extended thought processes have shown potential to significantly enhance reasoning performance, and reasoning distillation techniques enable smaller LLMs to produce extended thinking processes. This presents a valuable opportunity to investigate the reciprocal relationship between these capabilities. Our work provides a foundational exploration of this promising research direction, offering new insights into their interaction.

\section{Instances from DeepSeek-R1-Distill-XX Models (50 docs)}\label{appendix:instances}
\begin{figure*}[h]
    \centering
    \includegraphics[width=0.9\textwidth]{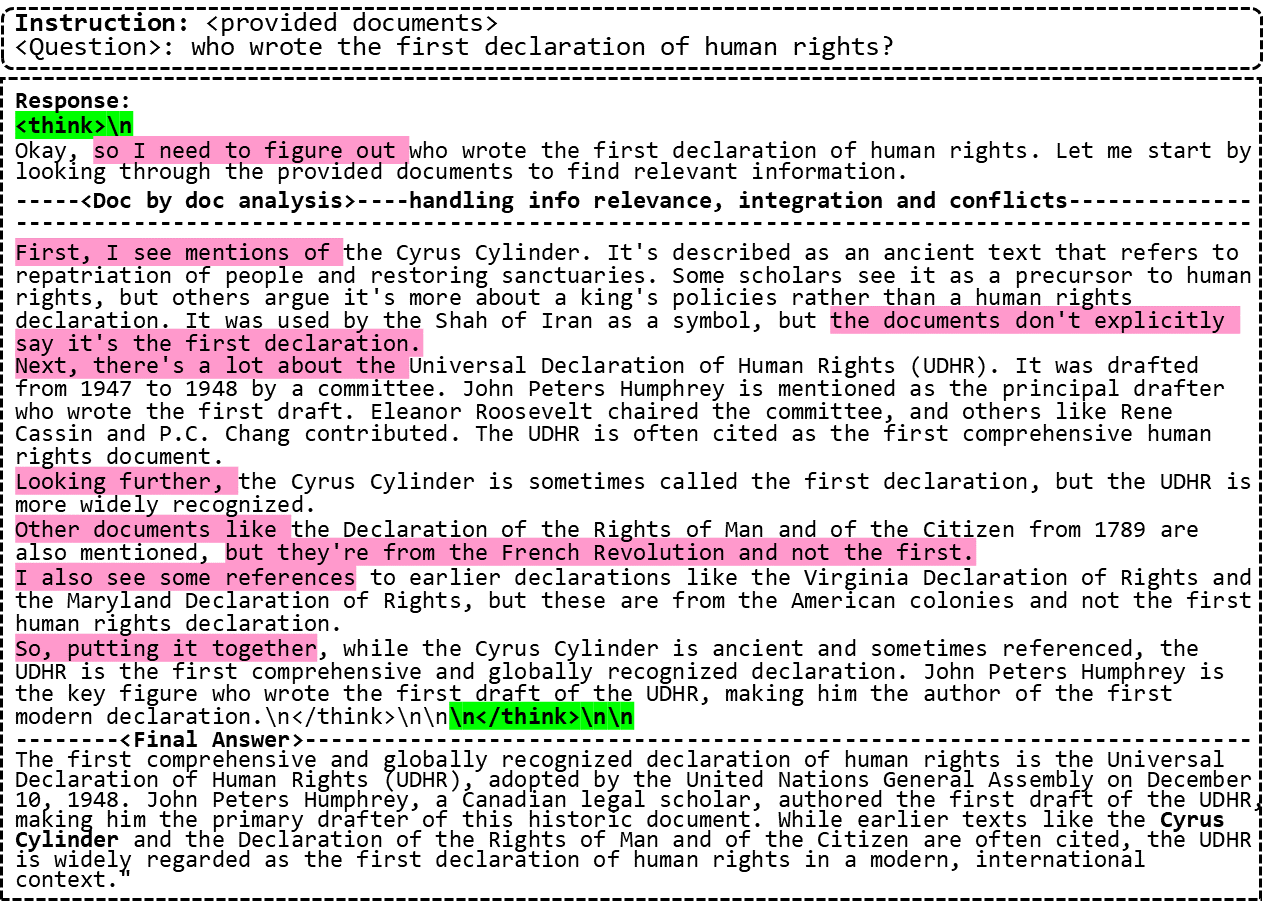}
    \caption{An instance of implementing long CoT on lengthy inputs including 50 docs from DeepSeek-R1-Distill-Qwen-14B.}
    \label{fig:enter-label}
\end{figure*}
\begin{figure*}[!h]
    \centering
    \includegraphics[width=0.9\textwidth]{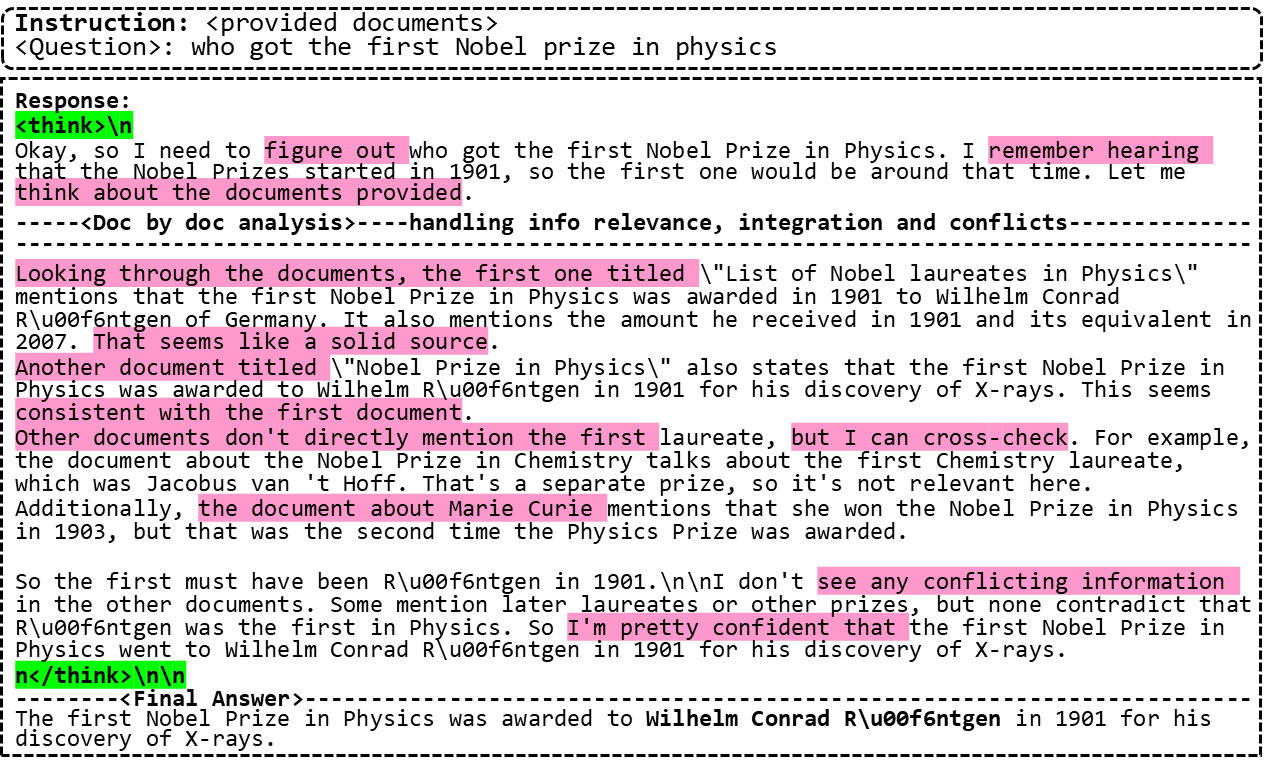}
    \caption{An instance of implementing long CoT on lengthy inputs including 50 docs from DeepSeek-R1-Distill-Qwen-32B.}
    \label{fig:enter-label}
\end{figure*}
\begin{figure*}[!h]
    \centering
    \includegraphics[width=0.9\textwidth]{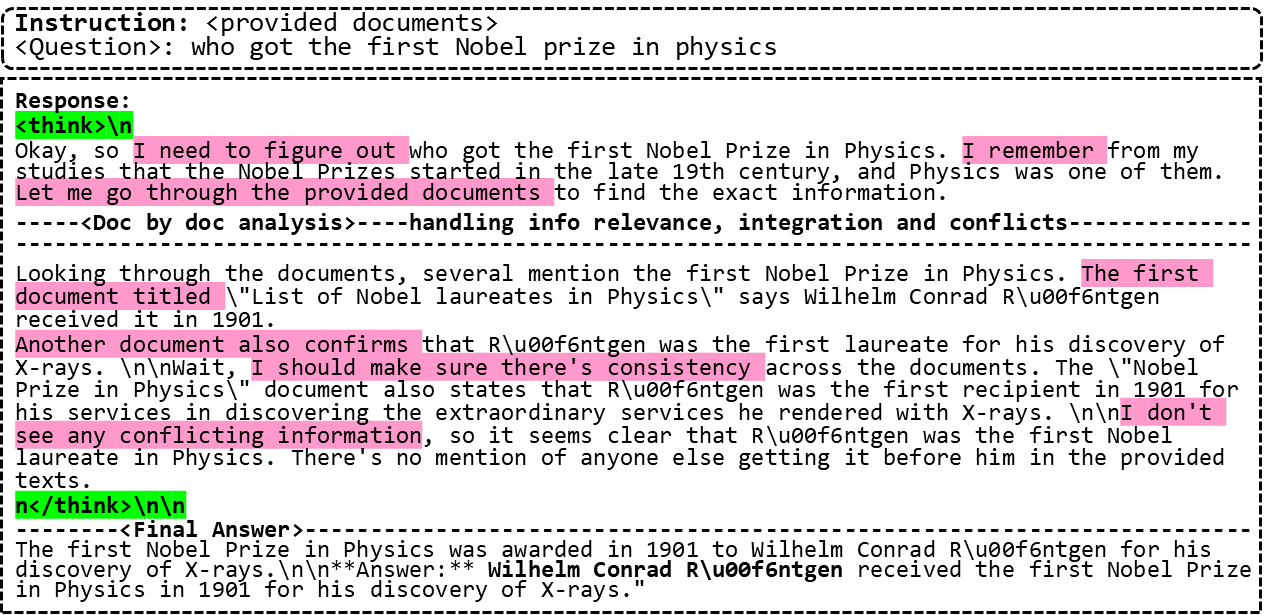}
    \caption{An instance of implementing long CoT on lengthy inputs from DeepSeek-R1-Distill-Llama-3.1-8B.}
    \label{fig:enter-label}
\end{figure*}

\begin{figure*}[!t]
    \centering
    \includegraphics[width=0.9\textwidth]{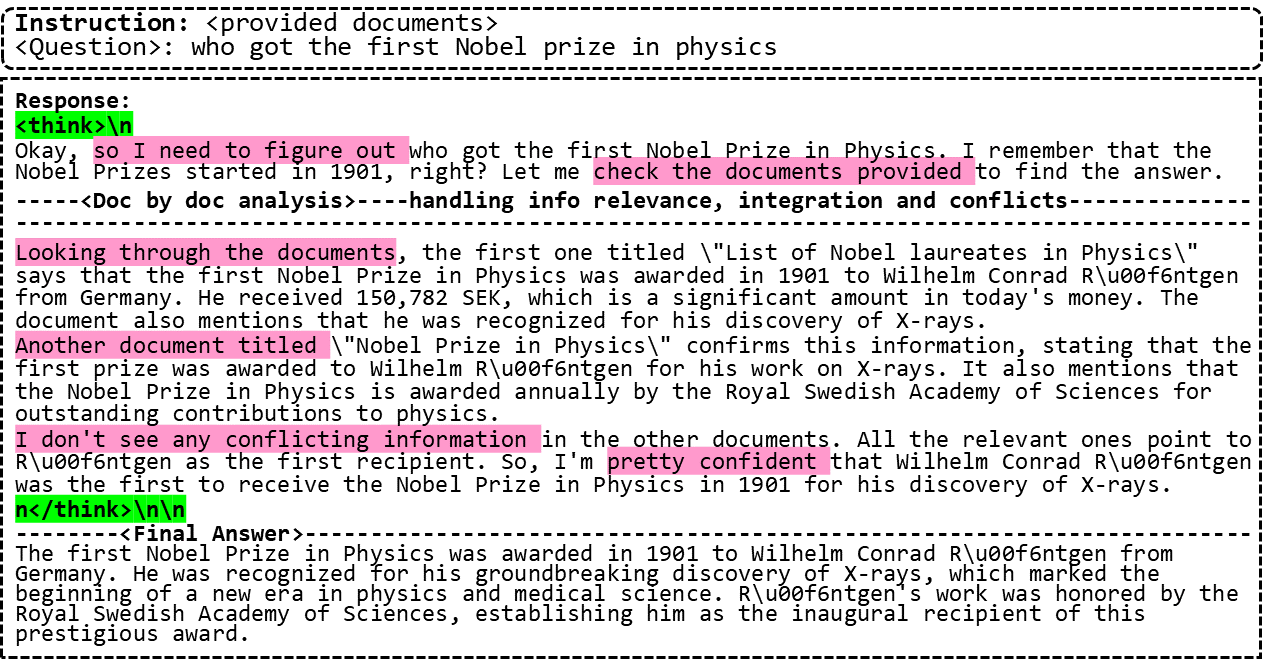}
    \caption{An instance of outputting long CoT on lengthy inputs  from DeepSeek-R1-Distill-Llama-3.3-70B-Instruct.}
    \label{fig:enter-label}
\end{figure*}
\vspace{-.2cm}
\section{Prompting Engineering}\label{appendix:prompt}
To effectively elicit the distilled reasoning patterns in distilled LLMs, we employ a specialized <think> token and utilize the DeepSeek-R1 chat template as the prompt framework for this series of distilled models. Notably, all instructional elements are systematically integrated within the user prompt structure. MDQA prompt is designed as follows:
\begin{figure*}[h]
    \centering
    \includegraphics[width=0.9\textwidth]{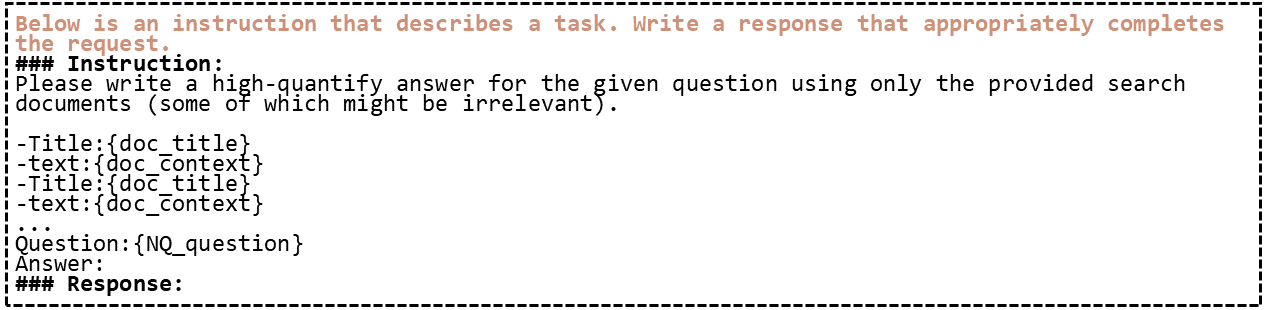}
    \caption{Prompt template design.}
    \label{fig:enter-label}
\end{figure*}
\end{document}